%% file: root.tex
\documentclass[letterpaper, 10 pt, conference]{ieeeconf}  % Comment this line out if you need a4paper

\IEEEoverridecommandlockouts                              % This command is only needed if 
\usepackage{cite}
\usepackage{amsmath,amssymb,amsfonts}
\usepackage{algorithm,algpseudocode}
\usepackage{relsize}
\usepackage{graphicx}
\usepackage{textcomp}
\usepackage{subcaption}
\usepackage{booktabs}
\usepackage[usenames,dvipsnames]{xcolor}
\usepackage{color}
\usepackage{dblfloatfix}
\graphicspath{{figs/}}
\usepackage{url}

\usepackage{tikz}
\usepackage{tikzit}
\input{fig1.tikzstyles}

\usepackage[colorlinks=true]{hyperref} 
\definecolor{mydarkblue}{rgb}{0,0.08,0.45}
\hypersetup{ 
	pdfborder=0 0 0,
	pdfpagemode=UseNone,
	colorlinks=true,
	linkcolor=mydarkblue,
	citecolor=mydarkblue,
	filecolor=mydarkblue,
	urlcolor=mydarkblue,
	pdfview=FitH}

\title{\LARGE \bf
Skeletal Feature Compensation for Imitation Learning with Embodiment Mismatch
}

\author{Eddy Hudson$^{1}$, Garrett Warnell$^{1,2}$, Faraz Torabi$^{1}$ and Peter Stone$^{1,3}$% <-this % stops a space
\thanks{$^1$\textit{The University of Texas at Austin} Austin, Texas, USA {\tt\small \{eddyhudson, faraztrb\}@utexas.edu, pstone@cs.utexas.edu}}
\thanks{$^2$\textit{Army Research Laboratory} Austin, Texas, USA {\tt\small garrett.a.warnell.civ@army.mil}}
\thanks{$^3$\textit{Sony AI} Austin, Texas, USA}
\thanks{This work has taken place in the Learning Agents Research
	Group (LARG) at UT Austin.  LARG research is supported in part by NSF
	(CPS-1739964, IIS-1724157, NRI-1925082), ONR (N00014-18-2243), FLI
	(RFP2-000), ARO (W911NF-19-2-0333), DARPA, Lockheed Martin, GM, and
	Bosch.  Peter Stone serves as the Executive Director of Sony AI
	America and receives financial compensation for this work.  The terms
	of this arrangement have been reviewed and approved by the University
	of Texas at Austin in accordance with its policy on objectivity in
	research.}}

\begin{document}

\maketitle
\thispagestyle{empty}
\pagestyle{empty}

%%%%%%%%%%%%%%%%%%%%%%%%%%%%%%%%%%%%%%%%%%%%%%%%%%%%%%%%%%%%%%%%%%%%%%%%%%%%%%%%

\begin{abstract}
	Learning from demonstrations in the wild (e.g. YouTube videos) is a tantalizing goal in imitation learning. However, for this goal to be achieved, imitation learning algorithms must deal with the fact that the demonstrators and learners may have bodies that differ from one another. This condition --- ``embodiment mismatch'' --- is ignored by many recent imitation learning algorithms. Our proposed imitation learning technique, SILEM (\textbf{S}keletal feature compensation for \textbf{I}mitation \textbf{L}earning with \textbf{E}mbodiment \textbf{M}ismatch), addresses a particular type of embodiment mismatch by introducing a learned affine transform to compensate for differences in the skeletal features obtained from the learner and expert. We create toy domains based on PyBullet's HalfCheetah and Ant to assess SILEM's benefits for this type of embodiment mismatch. We also provide qualitative and quantitative results on more realistic problems --- teaching simulated humanoid agents, including Atlas from Boston Dynamics, to walk by observing human demonstrations.
\end{abstract}

%%%%%%%%%%%%%%%%%%%%%%%%%%%%%%%%%%%%%%%%%%%%%%%%%%%%%%%%%%%%%%%%%%%%%%%%%%%%%%%%

\section{Introduction}

Endowing artificial agents with the ability to learn new behaviors by watching humans is a lofty goal in the field of machine learning.
Artificial agents that can learn in this way, i.e, perform {\em learning from demonstration} (LfD) or {\em imitation learning} (IL), have enormous learning potential.
First, IL enables machines to learn new skills from demonstrators that are experts in particular tasks of interest (e.g., operating construction equipment or cooking) rather than from researchers who are experts at writing computer code or designing cost functions.
Second, this paradigm of learning provides a channel of skill acquisition even in cases where it is currently prohibitively difficult to induce the desired behavior by writing computer code or specifying a cost function.
Finally, the greatest allure of being able to perform {\em LfD in the wild}---in which one seeks to enable an artificial agent to imitate new behaviors using unconstrained video demonstrations of those behaviors (e.g., YouTube videos)---is that there already exists a vast and relatively untapped amount of demonstration data that we can immediately use to drive behavior acquisition in machines.
Here, we are motivated by the problem of performing LfD in the wild for robot behavior acquisition, and we seek to make progress toward that goal in a very specific way.

While a great deal of recent research progress has been made on particular variants of imitation learning, relatively little recent work has considered the problem of {\em embodiment mismatch}, i.e., the situation that arises when a demonstrating agent's body is different than that of the imitator.
For example, a student in a fitness class who is 6'3'' is able to easily imitate a 5'2'' instructor doing jumping jacks.
While the specific signals sent to the student's muscles are much different, the jumping-jack motion induced is qualitatively the same.
For IL algorithms to be applied to unconstrained videos, they will similarly need to be able to deal with embodiment mismatch of this kind. We envision a two-stage pipeline where computer vision techniques such as keypoint extraction and pose-estimation are used to extract \emph{expert} data from unconstrained videos, and an IL algorithm capable of dealing with the embodiment mismatch in the expert data trains a learner to acquire the skills originally found in the video. 

In this paper, we focus our attention on the second stage of such a pipeline, and propose a new imitation learning algorithm called SILEM (\textbf{S}keletal feature compensation for \textbf{I}mitation \textbf{L}earning with \textbf{E}mbodiment \textbf{M}ismatch) that can help reliably train a learner to imitate an expert with a different body. Central to SILEM is a learned affine transform that compensates for differences in the skeletal features (e.g. joint angles, height of head, etc.) derived from the expert and learner.

What sets our work apart from prior work in this problem setting \cite{RoboImitationPeng20,peng2018sfv,gupta2017learning} is that 1) we do not require access to simulations of the expert's body, and 2) through a series of controlled ablation studies (Figures \ref{fig2}, \ref{fig3}), we provide empirical evidence of the detrimental effect embodiment mismatch has in the domains we consider before proceeding to address the issue.

\section{Background}

Our ultimate goal in this work is to learn a controller that solves a sequential decision making problem.
Such problems are typically formulated in the context of a Markov decision process (MDP), i.e., a tuple $\mathcal{M} = <\mathcal{S}, \mathcal{A}, T, R, \gamma>$, where $\mathcal{S}$ denotes an agent's state space, $\mathcal{A}$ denotes the agent's action space, $T: \mathcal{S} \times \mathcal{A} \rightarrow \Delta(\mathcal{S})$ denotes the environment model which maps state-action pairs to a distribution over the agent's next state, $R: \mathcal{S} \times \mathcal{A} \times \mathcal{S} \rightarrow \mathbb{R}$ is a reward function that provides a scalar-valued reward signal for state-action-next-state tuples, and $\gamma \in [0,1]$ is a discount factor that specifies how the agent should weight short- vs. long-term rewards.
Solutions to sequential decision making problems are often specified by reactive policies $\pi: \mathcal{S} \rightarrow \Delta(\mathcal{A})$, which specify agent behavior by providing a mapping from the agent's current state to a distribution over the actions it can take.
Machine learning solutions to problems described by an MDP typically search for policies that can maximize the agent's expected sum of future rewards.

The IL problem is typically formulated using an MDP {\em without} a specified reward function, i.e., $\mathcal{M}\setminus R$.
Instead of reward, the agent is provided with {\em demonstration} trajectories---typically assumed to have been generated by an expert---that specifies the desired behavior, i.e., $\tau_E = (s_0, a_0, s_1, a_1, ...)$. Imitation from observation (IfO) is a sub-problem of IL in which the agent does not have access to the actions taken during the demonstration trajectories, i.e., $\tau_E = (s_0, s_1, ...)$.
Techniques designed to solve the IL problem seek to use observed demonstrations to find policies that an imitating agent can use to imitate the demonstrator.

Adversarial imitation learning (AIL) is a particular way to perform IL that has recently come to the fore. AIL is loosely based on GANs \cite{goodfellow2014generative}, in that both involve the same min-max game with discriminators and generator networks. The discriminator $D$, is trained to distinguish between the demonstration trajectories and trajectories generated by the imitator.
In particular, the goal of updating $D$ is to drive $\mathbb{E}_{o\sim\tau_E}[D(o)]$ toward $1$ and $\mathbb{E}_{o\sim\tau}[D(o)]$ toward $0$, where $o$ is a segment of the trajectory, and $\tau$ represents trajectories recently generated by the imitator. In the seminal AIL algorithm GAIL \cite{ho2016generative}, $o=(s_t, a_t)$, whereas in GAIfO \cite{torabi2018generative}, $o=(s_t, s_{t+1}, \dots, s_{t+n})$. Merel \emph{et al.} \cite{merel2017learning} deviate from this paradigm by preventing the discriminator from accessing actual state information. Instead, they let $o=(g(s_t), g(s_{t+1}), \dots, g(s_{t+n}))$, where $g$ is an abstract function that extracts features based on the agent's state. In their IL experiments involving 2 and 3 link arms, $g$ simply returned the end effector's location. This representation allowed Merel \emph{et al.} \cite{merel2017learning} to train a 3-link arm to imitate a 2-link arm. In this work, we refer to the features extracted by $g$ as skeletal features. The generator, which in AIL algorithms is the imitator's policy $\pi$, is trained to induce behavior that elicits large output from $D$, i.e., to ``fool" $D$ into thinking that the imitator's trajectories came from the demonstrator.
By iteratively updating $D$ and $\pi$ as described, AIL approaches are able to find imitator policies that successfully mimic the demonstrated behavior. 

Key to this work is our observation that, to date, most AIL approaches that have been proposed require the expert and the learner to have the same embodiment. The embodiment mismatch problem we consider here occurs when the demonstrator is structurally different from the learner. For example, in the case of dog-like agents, the demonstrator may be a tall Great Dane and the learner may be an elongated Dachshund. In such cases, it might be impossible to match $\tau$ with $\tau_E$. AIL algorithms such as GAIL and GAIfO will tend to suffer degraded performance in this scenario. Therefore, solutions must be designed to explicitly account for this kind of embodiment mismatch. Note that our goal is not to train a single policy that is robust to changes in embodiment. Our aim is to train a policy specific to a particular embodiment by leveraging demonstrations from another embodiment.

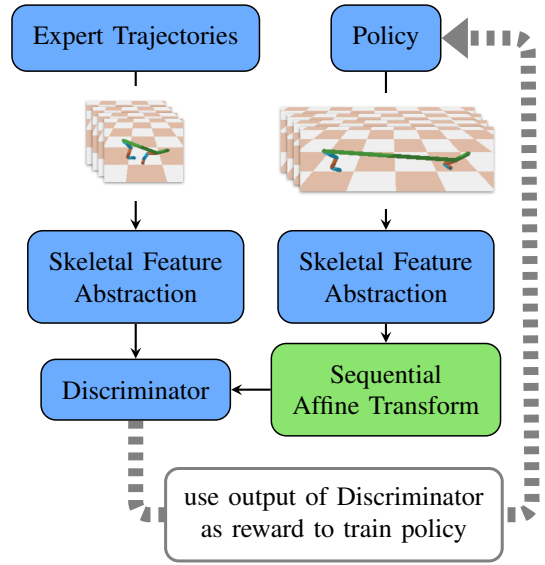
\begin{figure}[]
	\centering
	\input{figs/fig1a.tikz}
	\caption{A high level overview of the proposed technique, SILEM. The policy and discriminator are learned through a mini-max game as in prior AIL algorithms, while the sequential affine transform is learned by backpropagating through the discriminator (Algorithm \ref{alg:strail}).}
	\label{fig1}
\end{figure}

\section{SILEM}

In this work, we propose a new AIL algorithm capable of training useful policies in the presence of a particular type of embodiment mismatch, and term it SILEM (\textbf{S}keletal feature compensation for \textbf{I}mitation \textbf{L}earning with \textbf{E}mbodiment \textbf{M}ismatch). SILEM follows the lead of prior AIL algorithms GAIL and GAIfO in its core operations (Algorithm \ref{alg:strail}, Figure \ref{fig1}) with one critical exception --- the sequential affine transform. 

The sequential affine transform is an affine transform with the following functional form:

\begin{equation*}
\begin{split}
T(o_t) &= T\Big(g(s_t), g(s_{t+1}), \dots, g(s_{t+n})\Big) \\
&= \Big(f\big(g(s_t)\big), f\big(g(s_{t+1})\big), \dots, f\big(g(s_{t+n})\big)\Big)
\end{split}
\end{equation*}
\vspace{2pt}

\noindent where $T$ is the sequential affine transform, and $f$ is an affine transform operating on the skeletal features from single states. i.e., $f(g(s))=\textbf{A}g(s)+b$, where $\textbf{A}$ is a diagonal matrix and $g$ is an abstract function that extracts skeletal features from single states. We restrict our attention to IfO to more seamlessly learn from human demonstrations, which generally do not come with actions.

The sequential affine transform's purpose is to compensate for differences in the skeletal features obtained from expert and learner bodies, which arise due to embodiment mismatch. It is only applied to skeletal features from the learner, and is optimized by minimizing the following loss function: $L=-\log(D(T(o)))$, while keeping the weights of the discriminator $D$ fixed. $T$ is the sequential affine transform, and $o=(g(s_t), g(s_{t+1}), \dots, g(s_{t+n}))$ is a sequence of skeletal features from the learner. This loss function is similar to that employed for training Conditional GANs \cite{mirza2014conditional}. We frame the sequential affine transform as a conditional GAN that aims to generate skeletal features pertaining to the demonstrator given skeletal features from the learner. Combining this additional step with the essential components of GAIL/GAIfO, each iteration of SILEM comprises three main steps: (Line 9) update the discriminator $D$, (Line 10) update the sequential affine transform $T$, and (Line 11) update the policy $\pi$.

\begin{algorithm}[]
	\caption{SILEM and SILEM$^-$ (our ablation without the sequential affine transform). Lines in green to be executed only for SILEM (3, 7, 8, 10).}
	\label{alg:strail}
	\begin{algorithmic}[1]
		\State Initialize parametric policy $\pi$
		\State Initialize parametric discriminator $D$
		\State \textcolor{OliveGreen}{Initialize sequential affine transform $T$}
		\State Obtain state-only expert demonstration data $\tau_E=\{o_t\}=\{(s_t, s_{t+1}, \dots, s_{t+n})\}$
		\While{$\pi$ improves}
		\State Using $\pi$, collect learner trajectories $\tau=\{o_t\}=\{(s_t, s_{t+1}, \dots, s_{t+n})\}$
		\State \textcolor{OliveGreen}{Generate a copy of $\tau$ called $\tau_c$}
		\State \textcolor{OliveGreen}{Replace each element $o_t$ in $\tau$ with $T(o_t)$}
		\State Update $D$ using the loss: $-\bigg(\mathbb{E}_{o\sim\tau_E}[log(D(o))] + \mathbb{E}_{o\sim\tau}[log(1-D(o))]\bigg)$
		\State \textcolor{OliveGreen}{Update $T$ using the loss: $-\mathbb{E}_{o\sim\tau_c}[log(D(T(o)))]$}
		\State Update $\pi$ by performing PPO updates with reward function $D(o)$, where $o\in\tau$
		\EndWhile
	\end{algorithmic}
\end{algorithm}

Note that our choice of objective function to train the sequential affine transform runs the risk of interfering with policy learning. In the presence of embodiment mismatch, skeletal features from the learner can differ from the expert both due to embodiment mismatch and due to imperfect imitation. Left to its own devices, this objective function might encourage the sequential affine transform to go above and beyond its purpose by converting skeletal features from faltering learner states to skeletal features from excellent expert states -- as opposed to merely compensating for embodiment mismatch. However, we find empirically that, because we use an affine transform rather than a more powerful network (e.g., a multi-layer perceptron (MLP)), this scenario does not occur (Table \ref{tab1}).

At first glance, the restriction of having to use an affine transform might appear quite severe. It requires a certain degree of similarity between the expert and learner. However, as we show in the results section, by choosing $g$ appropriately, the space of problems solvable by SILEM widens to include potentially impactful problems such as training humanoid robots from human demonstrations. In our experiments involving human demonstrations, we define a separate $g$ for the demonstrator (\emph{expert}) and the learner, which allows us to address instances of embodiment mismatch where there exists a mismatch in the number of joints (see the humanoid walking experiments
in Section V). Naturally, as the difference between the expert and learner bodies becomes more and more drastic, the process for designing $g$ will become more and more involved. Fortuitously, as the expert differs from the learner in ever more ways, it becomes less and less useful to the learner since the feedback it can offer the learner also drops precipitously. For instance, humanoid robots can learn a lot more from human demonstrations than from demonstrations by quadrupedal agents.

\section{Related Work}

Our work broadly fits in the realm of learning from demonstrations (LfD), or imitation learning. In LfD, the learner is trained to imitate the behavior of an expert demonstrator. One use case for LfD is when it is difficult to specify the target behavior using a scalar reward signal. e.g. performing backflips \cite{christiano2017deep}. This is in contrast to many impressive results in the past decade using deep RL techniques such as in Atari games \cite{mnih2015human}, or Go \cite{silver2017mastering},  where the target behavior can be more easily achieved by specifying a scalar reward signal.

In LfD, the expert and learner might face different environments and have different embodiments. Prior work \cite{Liu2020State,Gangwani2020State-only} has shown promise in correcting the difference in dynamics faced by the learner and expert. However, these approaches have failed to show concrete evidence that they combat embodiment mismatch.

Manually matching the feature-space of the learner and expert can resolve embodiment mismatch to a certain extent \cite{merel2017learning,peng2018sfv}. However, methods such as these may be difficult to scale and often arrive at suboptimal solutions when the embodiment mismatch gets noticeably large. Peng \emph{et al.} \cite{peng2018sfv} also require control over the agent's starting state distribution, thus precluding application of their technique on real robots. Some methods for LfD sidestep the problem of constructing a mapping by using temporal consistency to learn a reward function that is invariant to the environment that the input state belongs to \cite{sermanet2018time}. However, Torabi \emph{et al.} \cite{torabi2018generative} show that it is difficult to learn periodic locomotion behaviors using such reward functions.

There have been a few attempts at automatically creating a mapping between the state-spaces of two different agents \cite{gupta2017learning,gamrian2018transfer,pmlr-v119-kim20c,RoboImitationPeng20,7803386}. Gupta \emph{et al.} \cite{gupta2017learning} require both the expert and the learner to have solved at least one common task beforehand. This makes it impossible to learn from new demonstrators in the wild, effectively ruling out any application for the LfD in the wild problem. Gamrian \emph{et al.} \cite{gamrian2018transfer} and Kim \emph{et al.} \cite{pmlr-v119-kim20c} use a framework based on GANs to map between various domains. However, they train the mapping using a random policy. In many robotics domains, random policies do not provide useful information: a random Humanoid agent will fall down immediately. Thus a mapping trained using such a policy is unlikely to generalize to walking gaits. Along those lines, Ariki \emph{et al.} \cite{7803386} also require random explorations, and they do show successful results using a humanoid robot (the NAO robot). However, they fail to evaluate their approach on more complex humanoid agents such as the ones we have used, where random policies fail to provide any useful information. Peng \emph{et al.} \cite{RoboImitationPeng20} use inverse kinematics to approximately match keypoints on the expert with keypoints on the learner. Such an approach, apart from having to be laboriously tuned, has only been shown to work for relatively simple examples of embodiment mismatch. Our best attempt to use the techniques of Peng \emph{et al.} \cite{RoboImitationPeng20} to train the humanoid agents to imitate human demonstrations fails to produce useful policies.

SILEM is closest to the work of Stadie \emph{et al.} \cite{Stadie2017ThirdPersonIL}, Third Person Imitation Learning (TPIL). TPIL employs one discriminator for policy improvement, and another separate discriminator to \emph{cancel} out embodiment mismatch using gradient reversal \cite{10.5555/3045118.3045244}. The success of the entire approach hinges on the hypothesis that differences in single states reflect only embodiment mismatch and not differences in policy quality. However, for complex domains such as humanoid walking, this hypothesis fails to hold \cite{merel2017learning}. Our results also confirm that TPIL fails to reliably help humanoid agents imitate human demonstrations.

Within the specific problem setting that we operate, learning locomotion skills in the presence of embodiment mismatch, we are the first to provide definitive evidence that our approach is capable of handling embodiment mismatch. Neither Peng {et al.} \cite{RoboImitationPeng20} nor Peng {et al.} \cite{peng2018sfv} perform controlled ablation studies showing that their approaches handle meaningful embodiment mismatch. Gangwani \emph{et al.} \cite{Gangwani2020State-only} perform such ablation studies, but they study transition dynamics mismatch. Ghadirzadeh \emph{et al.} \cite{DBLP:journals/corr/abs-2103-03697} and Yu \emph{et al.} \cite{DBLP:conf/rss/YuFDXZAL18} provide impressive results, but they do so on robotic arms, where the challenge is in dealing with a wide diversity of physical objects while the actual task remains simple: reaching and picking.

\begin{figure}[]
	\centering
	\begin{subfigure}[b]{0.4\textwidth}
		\hspace{-27pt}\input{figs/halfcheetah.tikz}
		\caption{The bodies we design based on HalfCheetah}
		\vspace{4pt}
		\label{fig2a}
	\end{subfigure}
	\begin{subfigure}[b]{0.4\textwidth}
		\centering
		\includegraphics[width=0.9\linewidth]{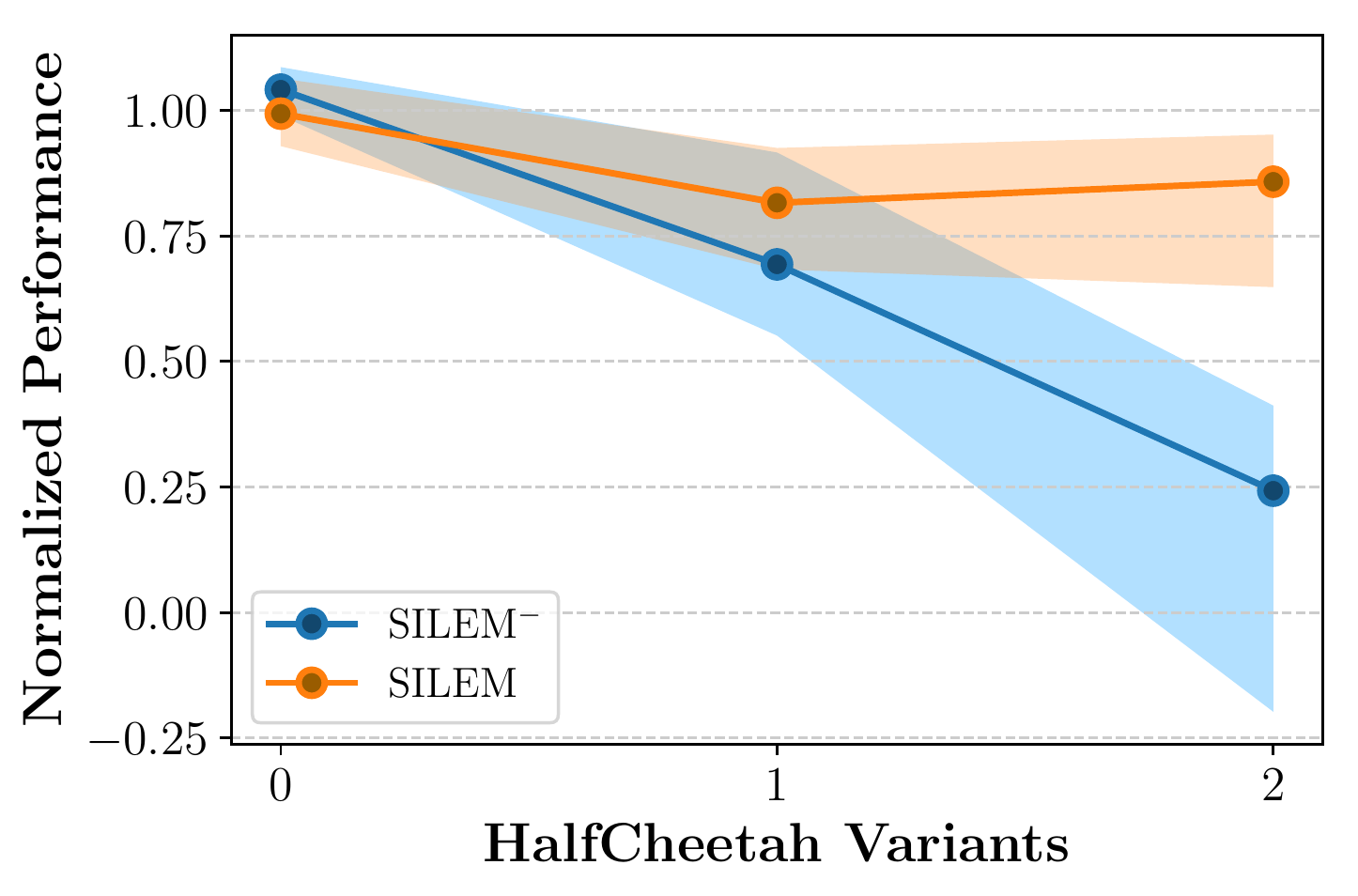}
		\caption{Results for the HalfCheetah bodies}
		\vspace{4pt}
		\label{fig2b}
	\end{subfigure}
	\begin{subfigure}[b]{0.4\textwidth}
		\hspace{-27pt}\input{figs/ant.tikz}
		\caption{The bodies we design based on Ant}
		\vspace{4pt}
		\label{fig2c}
	\end{subfigure}
	\begin{subfigure}[b]{0.4\textwidth}
		\centering
		\includegraphics[width=0.9\linewidth]{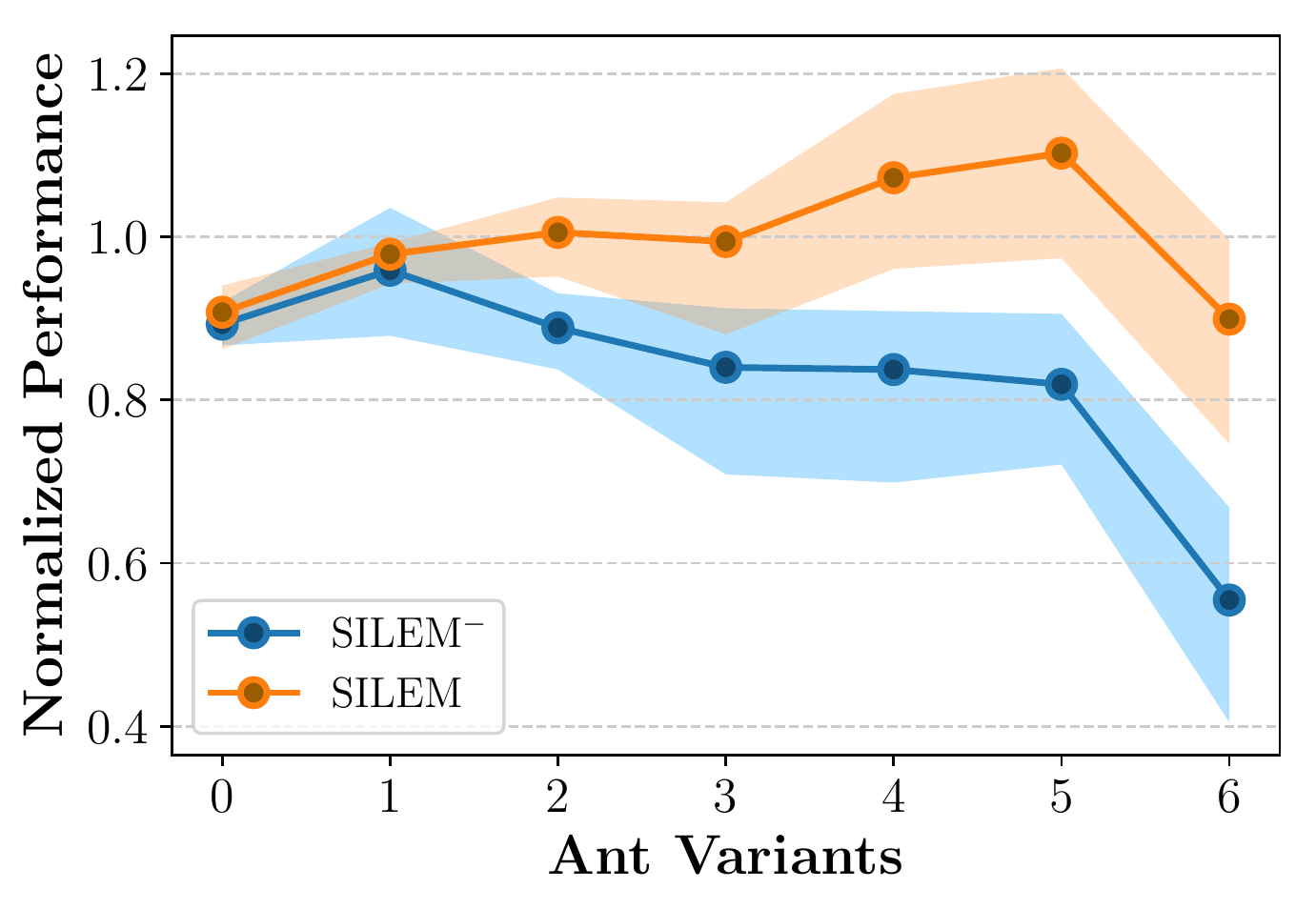}
		\caption{Results for the Ant bodies}
		\vspace{4pt}
		\label{fig2d}
	\end{subfigure}
	
	\caption{The toy domains we created to assess SILEM's benefits \textbf{(\ref{fig2b}, \ref{fig2d})} Results showing that SILEM is able to prevent degradation of imitation performance due to embodiment mismatch. SILEM$^-$ is an ablation of SILEM with the sequential affine transform removed from the training structure. The $y$-axis shows performance normalized by body-specific expert-level performance --- a score of 1.0 by an algorithm for a body indicates that that algorithm can match the performance of PPO on that body. The plot shows minimum, average, and maximum performance over 5 independent trials.}
	\label{fig2}
\end{figure}

\begin{figure*}[]
	\centering
	\hspace{11pt}\begin{subfigure}[b]{0.45\textwidth}
		\hspace{-33pt}\input{figs/humandemo.tikz}
		\caption{The simulated humanoids considered in this paper}
		\vspace{4pt}
		\label{fig3a}
	\end{subfigure}
	\begin{subfigure}[b]{0.45\textwidth}
		\hspace{33pt}\includegraphics[width=0.8\linewidth]{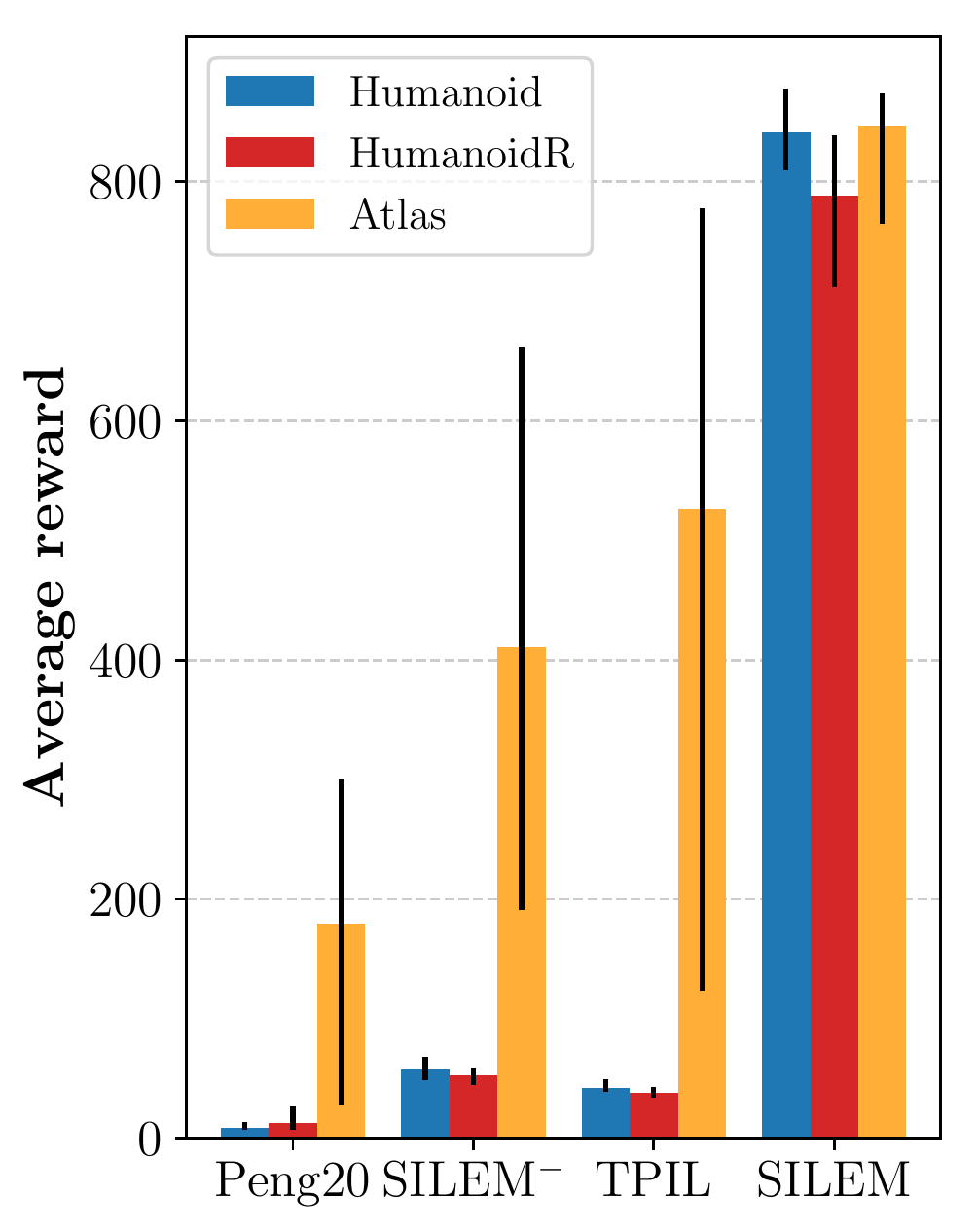}
		\caption{Results in Atlas, Humanoid, and HumanoidR}
		\vspace{4pt}
		\label{fig3b}
	\end{subfigure}
	\caption{The experiments we performed involving human demonstrations. \textbf{(\ref{fig3b})} Results showing that SILEM is able to reliably learn from human demonstrations, while TPIL, Peng20, and our ablation without the sequential affine transform (SILEM$^-$) are unable to do so. The $y$-axis uses the built-in reward function from the DeepMind Control Suite, which rewards the agent for maintaining an upright posture, matching the human's speed, and minimizing energy expenditure. The maximum reward possible is 1000. The plot shows minimum, average, and maximum performance over 5 independent trials.}
	\label{fig3}
\end{figure*}

\section{Experiments and Results}

Our experiments involve a series of HalfCheetah-based bodies (Figure \ref{fig2a}), a series of Ant-based bodies (Figure \ref{fig2c}), and three simulated humanoid agents Humanoid, HumanoidR, and Atlas (Figure \ref{fig3a}). The input to the policy network and $Q$ function consist of the default low level state information (for example joint angles, velocities, etc.). 

\begin{figure*}[]
	\centering
	\begin{subfigure}[b]{0.4\textwidth}
		\includegraphics[width=1\linewidth]{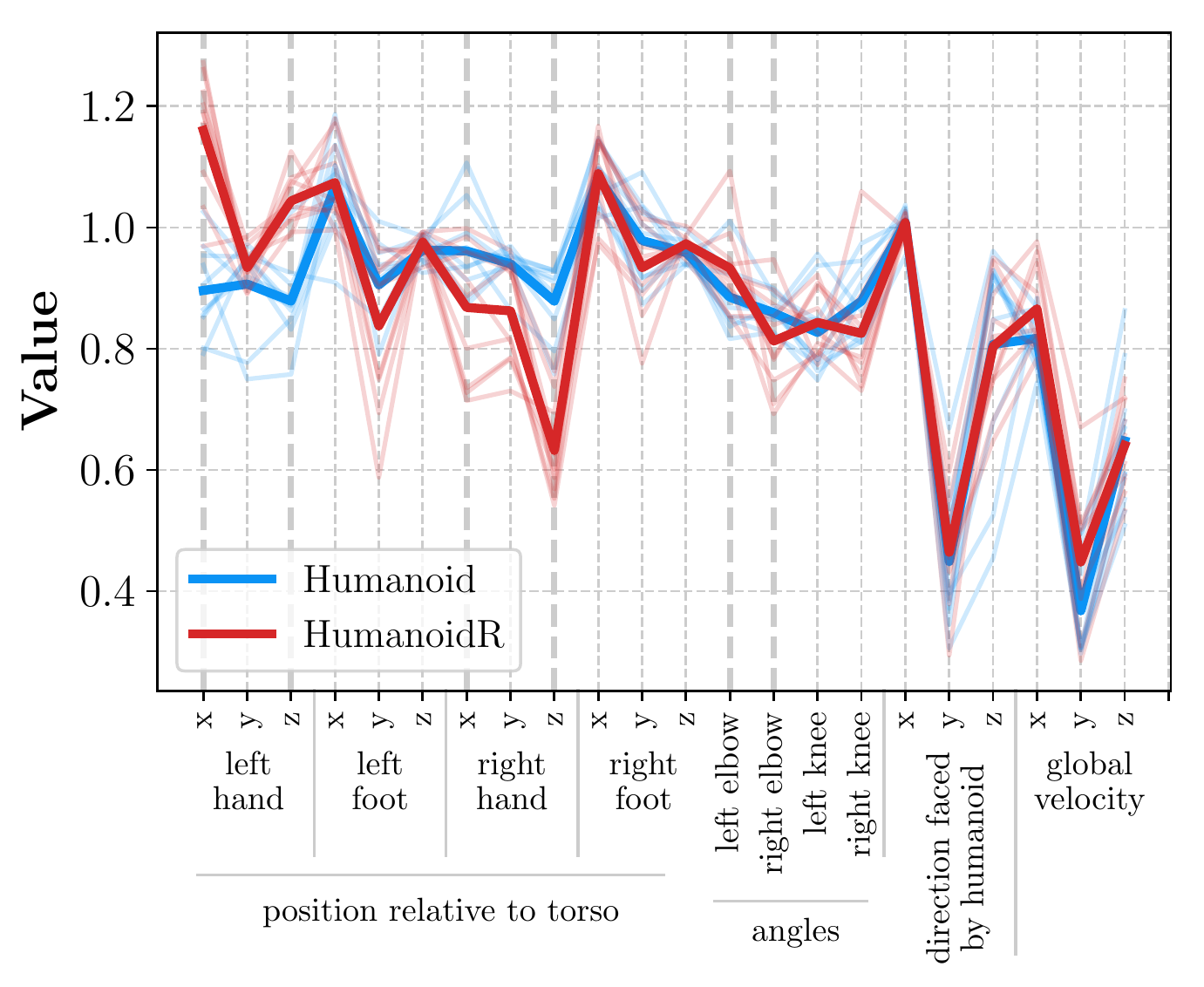}
		\caption{Values in the diagonal matrix, \textbf{A}}
		\vspace{4pt}
		\label{fig4a}
	\end{subfigure}
	\begin{subfigure}[b]{0.4\textwidth}
		\includegraphics[width=1\linewidth]{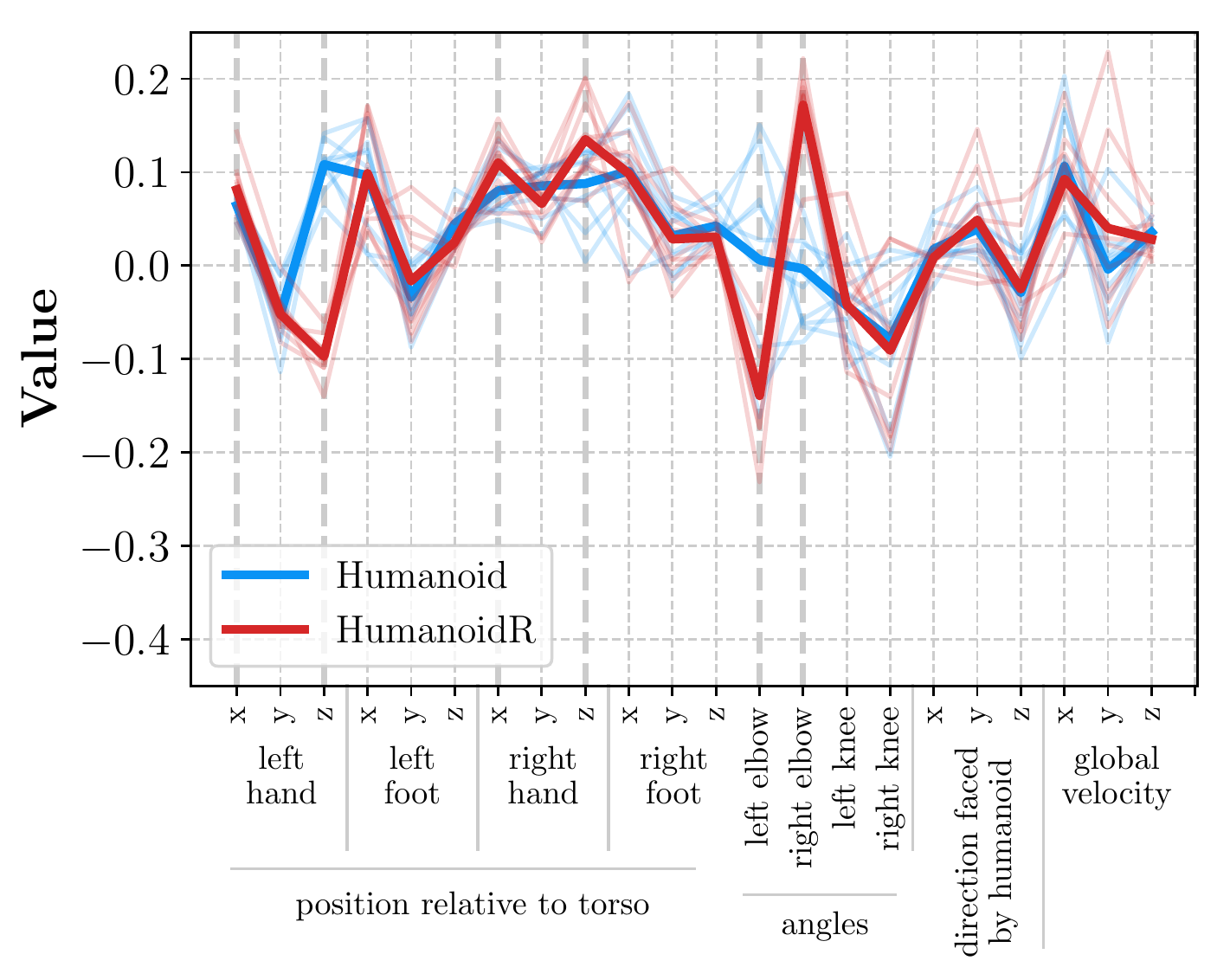}
		\vspace{-5pt}
		\caption{Values in the bias, $b$}
		\label{fig4b}
	\end{subfigure}
	\caption{The values of the affine transform from the top 10 policies in both Humanoid and HumanoidR. Thicker plotlines show the mean. The relevant features have been highlighted with thicker vertical gridlines. Note that positive $x$ faces forward from the agent, and positive $z$ points upwards. \textbf{(\ref{fig4a})} As expected, the relative position of the left and right hands get scaled up and down respectively (the hands' $z$-values are always negative). \textbf{(\ref{fig4b})} The left arm maintains a more obtuse angle than the right arm in order to match the counterbalance provided by the right arm (see supplementary video). The sequential affine transform allows for this adaptive correction by modifying the left and right elbow angles accordingly.}
	\label{fig4}
\end{figure*}

For the HalfCheetah-based and Ant-based bodies, $g$ is simply the identity function, while for the humanoid agents, it is a function that abstracts out the list of features defined in Figures \ref{fig4a} and \ref{fig4b}. In our experiments involving human demonstrations, there exists an asymmetry in the definition of $g$. For example, due to differences in the skeletal structure between the demonstrator and learner (Figure \ref{fig3a}), it is not possible to use the same function to compute the relative position of the right hand in the demonstrator and learner. Thus, we use different $g$ functions for the demonstrator and learner (that eventually end up computing the same quantity).

We generated the three HalfCheetah-based bodies HC0, HC1, and HC2 by setting the torso length at 1, 2, and 3 respectively. HC0 is the original HalfCheetah provided by PyBullet. We then train HC0, HC1, and HC2 to imitate an expert HC0 that is running as fast as possible. We generated the Ant-based bodies by increasing the length of the link closest to the original Ant's body in steps of 0.05 from 0.2 to 0.5. This resulted in a total of 7 Ant-based bodies, from Ant0 to Ant6, where Ant0 is the original Ant provided by PyBullet. We then train all the Ant-based bodies to imitate an expert Ant0 that is running as fast as possible. The first humanoid agent, Humanoid, is the simple humanoid from the DeepMind Control Suite \cite{tassa2018deepmind}. The second humanoid, HumanoidR, is an asymmetric humanoid that we generated by elongating Humanoid's right arm and shortening its left arm. The third humanoid is a simulated version of Atlas, a robot from Boston Dynamics.  The humanoid agents are trained to imitate the first demonstration from Subject 8 in the CMU Mocap Library (\href{http://mocap.cs.cmu.edu}{mocap.cs.cmu.edu}). The demonstration is of a human walking forward. Note that we do not train HumanoidR to imitate Humanoid. They both,
along with the simulated Atlas, are trained to imitate the human
demonstration.

\textbf{Can the sequential affine transform address embodiment mismatch?} To answer this question, we first create an ablation of SILEM, SILEM$^-$ (Algorithm \ref{alg:strail}), where the sequential affine transform is removed from the training structure. We then apply SILEM$^-$ and SILEM to the toy domains we designed and to train the three humanoids (Humanoid, HumanoidR, and Atlas). Our results in Figure \ref{fig2} and \ref{fig3b} and the attached supplementary video shows that SILEM is able to consistently generate stable gaits for all of the agents we test, while SILEM$^-$ fails to do so. 

As further evidence that the sequential affine transform addresses embodiment mismatch, we plot the affine transforms used by the best performing policies of Humanoid and HumanoidR in Figure \ref{fig4a} and \ref{fig4b}. The results show that the sequential affine transforms apply targeted compensation for the asymmetric arms, thus providing evidence that SILEM works by correcting embodiment mismatch.

\textbf{Comparison with TPIL and Peng20 on training humanoid agents from human demonstrations:} Note that we refer to our implementation of Peng {et al.} \cite{RoboImitationPeng20} as Peng20. We compare SILEM with Peng20 since our application of the sequential affine transform for humanoid agents is similar to their application of inverse kinematics for dog-like agents. Peng20 matches end effectors between demonstrator and learner, while we also additionally match the elbow and knee angles. We also compare SILEM to TPIL since we found TPIL to be the closest available alternative to SILEM --- like SILEM, TPIL is an AIL algorithm that addresses mismatch between expert and learner in an online fashion.

Our quantitative results in Figure \ref{fig3b} and our qualitative results in the attached supplementary video show that SILEM outperforms both Peng20 and TPIL by a large margin at the task of imitating the human demonstration. Surprisingly, Peng20, TPIL, and SILEM$^-$ perform much better on Atlas than they do on the other humanoid agents. We posit that this is because Atlas is closer in dimensions to the demonstrator than the other humanoid agents. The feet are 1.13, 1.26, and 0.99 units away from the torso for the demonstration, Humanoid, and Atlas respectively. However, the hands are 0.74, 0.4 and 0.91 units away from the torso for the demonstration, Humanoid, and Atlas respectively.

\textbf{Why does the sequential affine transform not interfere with the policy's learning?} The objective function used to learn the sequential affine transform has the potential to conflict with the policy's learning objective --- the sequential affine transform could, in theory, generate skeletal features corresponding to good expert states from skeletal features corresponding to poor learner states. We hypothesize that such a scenario does not arise because the sequential affine transform consists of affine transforms and not more powerful deep networks such as MLPs. Our experiments support the claim (Table \ref{tab1}). For context, the average ground truth reward obtained by a random policy and expert is -1555 and 2249 respectively. On the other hand, the discriminator reward can vary from 0 to 1000. Using an MLP (1 hidden layer with 100 units) results in high discriminator reward, but poor performance by the policy. This finding suggests that the MLP is crafting skeletal features corresponding to good expert states from skeletal features corresponding to poor learner states.

\begin{table}[h]
	\centering
	\caption{Results from applying SILEM to HC0 where $f$ in the sequential affine transform is either an MLP or an Affine Transform. These results suggest a degeneracy in the MLP not present with the affine transform --- the MLP is both compensating for embodiment mismatch and imperfect imitation. All results are the mean over 5 independent trials.}
	\begin{tabular}{@{}ccc@{}}
		\toprule
		$f$                & \textbf{\begin{tabular}[c]{@{}c@{}}Discriminator\\ Reward\end{tabular}} & \textbf{\begin{tabular}[c]{@{}c@{}}Ground Truth\\ Reward\end{tabular}} \\ \midrule
		MLP              & 466                                                                     & -1979                                                                  \\
		Affine Transform & 451                                                                     & 2234                                                                   \\ \bottomrule
	\end{tabular}
	\label{tab1}
\end{table}

\section{Conclusion}

In this paper, we have introduced a new AIL algorithm called SILEM (\textbf{S}keletal feature compensation for \textbf{I}mitation \textbf{L}earning with \textbf{E}mbodiment \textbf{M}ismatch). Using SILEM, we showed improved performance in the challenging problem of learning from human demonstrations. We presented evidence that SILEM works by learning a sequential affine transform capable of compensating for differences in the skeletal features of the expert and learner that arise due to embodiment mismatch. 

We hypothesize that SILEM is also capable of handling simple cases of \emph{environment} mismatch. E.g., the human walking demonstration is from a flat surface, but the humanoid agent is learning to walk on an inclined surface. By scaling and shifting the skeletal features appropriately, it might be possible to make the walking behavior on an inclined surface resemble that on a flat surface. An interesting line of future work would be to exhaustively evaluate the capabilities of SILEM with regards to environment mismatch. This could be followed by trying to successfully replace the sequential affine transform with a more sophisticated architecture, such as an ensemble of affine transforms, thus opening the door for SILEM to tackle ever more complex instances of embodiment and environment mismatch.

\end{document}

\maketitle

\section{Hyper-parameter search}

For every experiment, we used a grid search to find the best set of hyper-parameters. Specifically, for every configuration of hyper-parameters in the experiment, we ran 5 independent trials. Within each trial, we trained a policy, measured the average reward obtained by that policy over 1000 episodes, and assigned the resulting number as the score for that particular trial. The score for each configuration of hyper-parameters is the average score over the 5 trials corresponding to that configuration. The best configuration of hyper-parameters is then that which maximizes this score. 

All instantiations of SILEM, SILEM $^-$, and TPIL, use $n=4$. I.e., the input to the discriminator is a tuple of size 4. 

\subsection{General settings for PPO}

Throughout this work, the learning rate in PPO is always linearly decayed to $0$. We use the Adam optimizer with the default options from OpenAI baselines. The other settings in Table \ref{tab2} and Table \ref{tab2_h} are also used throughout this work unless specified otherwise. Note that both SILEM and SILEM$^-$ use PPO to improve the policy, but instead of the reward function from the environment, they utilize the output of the discriminator as the reward function.  Contrary to Table \ref{tab2}, the Ant experts were trained for 1000 PPO iterations.

\begin{table}[h!]
	\vskip 0.15in
	\begin{center}
		\parbox{11cm}{\caption{Common hyper-parameters used for PPO in experiments involving Atlas, Humanoid and HumanoidR (The Mujoco environment).}\label{tab2_h}}
		\vspace{11pt}
		
		\begin{small}
			\begin{sc}
				\begin{tabular}{lcccr}
					\toprule
					Hyper-parameter & Values \\
					\midrule
					No. timesteps used for each iteration    & 4096 \\
					PPO clip parameter & 0.1\\
					discount factor ($\gamma$)    & 0.99 \\
					lambda ($\lambda$)    & 0.95 \\
					PPO minibatch size       & 256 \\
					No. PPO iterations       & 8000 \\
					Learning rate       & 1e-4 \\
					\bottomrule
				\end{tabular}
			\end{sc}
		\end{small}
	\end{center}
	\vskip -0.1in
\end{table}
\newpage

\begin{table}[h!]
	\vskip 0.15in
	\begin{center}
		\parbox{11cm}{\caption{Common hyper-parameters used for PPO in experiments involving Ant and HalfCheetah (The Pybullet environments).}\label{tab2}}
		\vspace{11pt}
		
		\begin{small}
			\begin{sc}
				\begin{tabular}{lcccr}
					\toprule
					Hyper-parameter & Values \\
					\midrule
					No. timesteps used for each iteration    & 2048 \\
					PPO clip parameter & 0.2\\
					discount factor ($\gamma$)    & 0.99 \\
					lambda ($\lambda$)    & 0.95 \\
					No. PPO epochs      & 10 \\
					PPO minibatch size       & 64 \\
					No. PPO iterations       & 2000 \\
					Learning rate       & 2e-4 \\
					\bottomrule
				\end{tabular}
			\end{sc}
		\end{small}
	\end{center}
	\vskip -0.1in
\end{table}

\subsection{SILEM and SILEM$^-$ for the humanoid agents (Atlas, Humanoid and HumanoidR)}

The SILEM experiments involving humanoid agents used L2 normalization for the state transformer's weights. None of our other SILEM experiments used L2 normalization. The diagonal matrix and bias values of the state transformer were normalized with different coefficients.

\begin{table}[h!]
	\label{tab4_h}
	\vskip 0.15in
	\begin{center}
		\parbox{11cm}{\caption{The grid search for SILEM and SILEM$^-$ on Humanoid and HumanoidR was conducted by evaluating all permutations of the following parameter values. Hyper-parameters in bold only apply to SILEM.}}
		\vspace{11pt}
		
		\begin{small}
			\begin{sc}
				\begin{tabular}{lcccr}
					\toprule
					Hyper-parameter & Values \\
					\midrule
					($N_D$) No. update steps for the discriminator & 1, 2\\
					($N_G$) \textbf{No. update steps for the state transformer}    & 1 \\
					($d$) \textbf{Learning rate decay for the state transformer}    & 0.996, 1 \\
					($w_d$) \textbf{L2 coefficient for diagonal matrix}    & 1, 10 \\
					($w_b$) \textbf{L2 coefficient for  bias values}    & 1, 10 \\
					($N_P$) No. PPO epochs      & 4, 7 \\
					\bottomrule
				\end{tabular}
			\end{sc}
		\end{small}
	\end{center}
	\vskip -0.1in
\end{table}

\begin{table}[h!]
	
	\label{tab4.1_h}
	\vskip 0.15in
	\begin{center}
		\parbox{11cm}{\caption{The combination of parameters that turned out best for each experiment involving SILEM and Humanoid and HumanoidR.}}
		\vspace{11pt}
		
		\begin{small}
			\begin{sc}
				\begin{tabular}{lccccc}
					\toprule
					humanoid agent & $N_D$ & $d$ & $w_d$ & $w_b$ & $N_P$ \\
					\midrule
					Humanoid    & 2 & 1 & 10 & 10 & 7\\
					HumanoidR & 1 & 1 & 1 & 10 & 7\\
					\bottomrule
				\end{tabular}
			\end{sc}
		\end{small}
	\end{center}
	\vskip -0.1in
\end{table}

\begin{table}[h!]
	
	\label{tab4.2_h}
	\vskip 0.15in
	\begin{center}
		\parbox{11cm}{\caption{The combination of parameters that turned out best for each experiment involving SILEM$^-$ and Humanoid and HumanoidR.}}
		\vspace{11pt}
		
		\begin{small}
			\begin{sc}
				\begin{tabular}{lccccc}
					\toprule
					humanoid agent & $N_D$ & $N_P$ \\
					\midrule
					Humanoid    & 1 & 7\\
					HumanoidR & 1 & 7\\
					\bottomrule
				\end{tabular}
			\end{sc}
		\end{small}
	\end{center}
	\vskip -0.1in
\end{table}

\begin{table}[h!]
	\label{tab42_h}
	\vskip 0.15in
	\begin{center}
		\parbox{11cm}{\caption{The grid search for SILEM and SILEM$^-$ on Atlas was conducted by evaluating all permutations of the following parameter values. Hyper-parameters in bold only apply to SILEM.}}
		\vspace{11pt}
		
		\begin{small}
			\begin{sc}
				\begin{tabular}{lcccr}
					\toprule
					Hyper-parameter & Values \\
					\midrule
					($N_D$) No. update steps for the discriminator & 1, 2\\
					($N_G$) \textbf{No. update steps for the state transformer}    & 2, 3 \\
					($w_d$) \textbf{L2 coefficient for diagonal matrix}    & 1, 10 \\
					($w_b$) \textbf{L2 coefficient for  bias values}    & 1, 10 \\
					($N_P$) No. PPO epochs      & 5, 8, 12 \\
					($a$) Ankle's range of motion      & $\pm0.3$, $\pm0.1$ \\
					\bottomrule
				\end{tabular}
			\end{sc}
		\end{small}
	\end{center}
	\vskip -0.1in
\end{table}

\begin{table}[h!]
	
	\label{tab42.1_h}
	\vskip 0.15in
	\begin{center}
		\parbox{11cm}{\caption{The combination of parameters that turned out best for each experiment involving SILEM and Atlas.}}
		\vspace{11pt}
		
		\begin{small}
			\begin{sc}
				\begin{tabular}{lcccr}
					\toprule
					Hyper-parameter & Values \\
					\midrule
					($N_D$) No. update steps for the discriminator & 2\\
					($N_G$) {No. update steps for the state transformer}    & 2 \\
					($w_d$) {L2 coefficient for diagonal matrix}    & 10 \\
					($w_b$) {L2 coefficient for  bias values}    & 10 \\
					($N_P$) No. PPO epochs      & 8 \\
					($a$) Ankle's range of motion      & $\pm0.1$ \\
					\bottomrule
				\end{tabular}
			\end{sc}
		\end{small}
	\end{center}
	\vskip -0.1in
\end{table}

\begin{table}[h!]
	
	\label{tab42.2_h}
	\vskip 0.15in
	\begin{center}
		\parbox{11cm}{\caption{The combination of parameters that turned out best for each experiment involving SILEM$^-$ and Atlas.}}
		\vspace{11pt}
		
		\begin{small}
			\begin{sc}
				\begin{tabular}{lcccr}
					\toprule
					Hyper-parameter & Values \\
					\midrule
					($N_D$) No. update steps for the discriminator & 1\\
					($N_P$) No. PPO epochs      & 8 \\
					($a$) Ankle's range of motion      & $\pm0.1$ \\
					\bottomrule
				\end{tabular}
			\end{sc}
		\end{small}
	\end{center}
	\vskip -0.1in
\end{table}

\newpage

\subsection{TPIL (Third Person Imitation Learning) for Atlas, Humanoid and HumanoidR}

\begin{table}[h!]
	\label{tab4.3_h}
	\vskip 0.15in
	\begin{center}
		\parbox{11cm}{\caption{The grid search for TPIL on Humanoid and HumanoidR was conducted by evaluating all permutations of the following parameter values.}}
		\vspace{11pt}
		
		\begin{small}
			\begin{sc}
				\begin{tabular}{lcccr}
					\toprule
					Hyper-parameter & Values \\
					\midrule
					($C_g$) Gradient reversal coefficient & 0.1, 0.3, 0.6, 1, 1.3, 1.6\\
					($N_D$) No. update steps for the discriminator & 1, 2, 3\\
					($N_P$) No. PPO epochs      & 4, 5, 7 \\
					\bottomrule
				\end{tabular}
			\end{sc}
		\end{small}
	\end{center}
	\vskip -0.1in
\end{table}

\newpage

\begin{table}[h!]
	
	\label{tab4.4_h}
	\vskip 0.15in
	\begin{center}
		\parbox{11cm}{\caption{The combination of parameters that turned out best for each experiment involving TPIL and Humanoid and HumanoidR.}}
		\vspace{11pt}
		
		\begin{small}
			\begin{sc}
				\begin{tabular}{lccccc}
					\toprule
					humanoid agent & $C_g$ & $N_D$ & $N_P$ \\
					\midrule
					Humanoid    & 0.6 & 1 & 5\\
					HumanoidR & 0.3 & 1 & 4\\
					\bottomrule
				\end{tabular}
			\end{sc}
		\end{small}
	\end{center}
	\vskip -0.1in
\end{table}

\begin{table}[h!]
	\label{tab42.3_h}
	\vskip 0.15in
	\begin{center}
		\parbox{11cm}{\caption{The grid search for TPIL on Atlas was conducted by evaluating all permutations of the following parameter values.}}
		\vspace{11pt}
		
		\begin{small}
			\begin{sc}
				\begin{tabular}{lcccr}
					\toprule
					Hyper-parameter & Values \\
					\midrule
					($C_g$) Gradient reversal coefficient & 0.1, 0.3, 0.6, 1, 1.3, 1.6\\
					($N_D$) No. update steps for the discriminator & 1, 2\\
					($N_P$) No. PPO epochs      & 5, 8, 12 \\
					($a$) Ankle's range of motion      & $\pm0.1$, $\pm0.3$ \\
					\bottomrule
				\end{tabular}
			\end{sc}
		\end{small}
	\end{center}
	\vskip -0.1in
\end{table}

\begin{table}[h!]
	
	\label{tab42.4_h}
	\vskip 0.15in
	\begin{center}
		\parbox{11cm}{\caption{The combination of parameters that turned out best for each experiment involving TPIL and Atlas.}}
		\vspace{11pt}
		
		\begin{small}
			\begin{sc}
				\begin{tabular}{lcccr}
					\toprule
					Hyper-parameter & Values \\
					\midrule
					($C_g$) Gradient reversal coefficient & 1.3\\
					($N_D$) No. update steps for the discriminator & 1\\
					($N_P$) No. PPO epochs      & 8 \\
					($a$) Ankle's range of motion      & $\pm0.1$ \\
					\bottomrule
				\end{tabular}
			\end{sc}
		\end{small}
	\end{center}
	\vskip -0.1in
\end{table}

\subsection{Peng20 for Atlas, Humanoid and HumanoidR}

We first used \texttt{ik\_humanoid.py} to convert the demonstration trajectory to Humanoid's and HumanoidR's embodiment. Then we used \texttt{ik\_atlas.py}, to get the demonstration trajectory in Atlas' embodiment. Both the aforementioned python files can be found in the source code. With the demonstration trajectory in the correct embodiments, we now apply GAIfO/SILEM$^-$ to train the respective humanoid agents. We use the same grid search options as in Table \ref{tab4_h} and \ref{tab42_h}.

\begin{table}[h!]
	
	\label{tab421.4_h}
	\vskip 0.15in
	\begin{center}
		\parbox{11cm}{\caption{The combination of parameters that turned out best for each experiment involving Peng20 and Atlas.}}
		\vspace{11pt}
		
		\begin{small}
			\begin{sc}
				\begin{tabular}{lcccr}
					\toprule
					Hyper-parameter & Values \\
					\midrule
					($N_D$) No. update steps for the discriminator & 1\\
					($N_P$) No. PPO epochs      & 12 \\
					($a$) Ankle's range of motion      & $\pm0.3$ \\
					\bottomrule
				\end{tabular}
			\end{sc}
		\end{small}
	\end{center}
	\vskip -0.1in
\end{table}

\begin{table}[h!]
	
	\label{tab4.4_h}
	\vskip 0.15in
	\begin{center}
		\parbox{11cm}{\caption{The combination of parameters that turned out best for each experiment involving Peng20 and Humanoid and HumanoidR.}}
		\vspace{11pt}
		
		\begin{small}
			\begin{sc}
				\begin{tabular}{lccccc}
					\toprule
					humanoid agent  & $N_D$ & $N_P$ \\
					\midrule
					Humanoid    & 1 & 7\\
					HumanoidR  & 1 & 4\\
					\bottomrule
				\end{tabular}
			\end{sc}
		\end{small}
	\end{center}
	\vskip -0.1in
\end{table}

\newpage

\subsection{SILEM and SILEM$^-$ for the HalfCheetah-based bodies}

\begin{table}[h!]
	\label{tab4}
	\vskip 0.15in
	\begin{center}
		\parbox{11cm}{\caption{The grid search for SILEM and SILEM$^-$ was conducted by evaluating all permutations of the following parameter values. Hyper-parameters in bold only apply to SILEM.}}
		\vspace{11pt}
		
		\begin{small}
			\begin{sc}
				\begin{tabular}{lcccr}
					\toprule
					Hyper-parameter & Values \\
					\midrule
					($n$) No. expert trajectories    & 10, 20 \\
					($N_D$) No. update steps for the discriminator & 5, 10\\
					($N_G$) \textbf{No. update steps for the state transformer}    & 5, 10 \\
					($d$) \textbf{Learning rate decay for the state transformer}    & 0.999, 1 \\
					($N_P$) No. PPO epochs      & 5, 10 \\
					($N_{Pmb}$) PPO minibatch size       & 64, 128 \\
					\bottomrule
				\end{tabular}
			\end{sc}
		\end{small}
	\end{center}
	\vskip -0.1in
\end{table}

\begin{table}[h!]
	
	\label{tab4.1}
	\vskip 0.15in
	\begin{center}
		\parbox{11cm}{\caption{The combination of parameters that turned out best for each experiment involving SILEM and a HalfCheetah-based body. We only ran SILEM using the HalfCheetah0 expert.}}
		\vspace{11pt}
		
		\begin{small}
			\begin{sc}
				\begin{tabular}{lcccccc}
					\toprule
					HalfCheetah-based body & $n$ & $N_D$ & $N_G$ & $d$ & $N_P$ & $N_{Pmb}$ \\
					\midrule
					HalfCheetah0    & 20 & 10 & 5 & 0.999 & 10 & 64\\
					HalfCheetah1 & 20 & 10 & 5 & 0.999 & 10 & 64\\
					HalfCheetah2    & 10  & 10 & 10 & 1 & 10 & 64\\
					\bottomrule
				\end{tabular}
			\end{sc}
		\end{small}
	\end{center}
	\vskip -0.1in
\end{table}

\begin{table}[h!]
	
	\label{tab4.2}
	\vskip 0.15in
	\begin{center}
		\parbox{11cm}{\caption{The combination of parameters that turned out best for each experiment involving SILEM$^-$ and a HalfCheetah-based body, when the HalfCheetah0 expert is used.}}
		\vspace{11pt}
		
		\begin{small}
			\begin{sc}
				\begin{tabular}{lcccccc}
					\toprule
					HalfCheetah-based body & $n$ & $N_D$ & $N_P$ & $N_{Pmb}$ \\
					\midrule
					HalfCheetah0    & 20 & 10 & 10 & 64\\
					HalfCheetah1 & 20 & 5 & 10 & 256\\
					HalfCheetah2    & 20  & 5 & 5 & 64\\
					\bottomrule
				\end{tabular}
			\end{sc}
		\end{small}
	\end{center}
	\vskip -0.1in
\end{table}

\newpage

\begin{table}[h!]
	
	\label{tab4.3}
	\vskip 0.15in
	\begin{center}
		\parbox{11cm}{\caption{The combination of parameters that turned out best for each experiment involving SILEM$^-$ and a HalfCheetah-based body, when the learner and expert have the same HalfCheetah-based body.}}
		\vspace{11pt}
		
		\begin{small}
			\begin{sc}
				\begin{tabular}{lcccccc}
					\toprule
					HalfCheetah-based body & $n$ & $N_D$ & $N_P$ & $N_{Pmb}$ \\
					\midrule
					HalfCheetah0    & 10 & 10 & 10 & 128\\
					HalfCheetah1 & 20 & 5 & 5 & 64\\
					HalfCheetah2    & 20  & 5 & 10 & 128\\
					\bottomrule
				\end{tabular}
			\end{sc}
		\end{small}
	\end{center}
	\vskip -0.1in
\end{table}

\subsection{SILEM and SILEM$^-$ for the Ant-based bodies}

\begin{table}[h!]
	
	\label{tab3}
	\vskip 0.15in
	\begin{center}
		\parbox{11cm}{\caption{The grid search for SILEM and SILEM$^-$ was conducted by evaluating all permutations of the following parameter values. Hyper-parameters in bold only apply to SILEM.}}
		\vspace{11pt}
		
		\begin{small}
			\begin{sc}
				\begin{tabular}{lcccr}
					\toprule
					Hyper-parameter & Values \\
					\midrule
					($n$) No. expert trajectories    & 10, 20 \\
					($N_D$) No. update steps for the discriminator & 5, 10\\
					($N_G$) \textbf{No. update steps for the state transformer}    & 5, 10 \\
					($d$) \textbf{Learning rate decay for the state transformer}    & 0.994, 0.996, 0.998, 0.999, 1 \\
					($N_P$) No. PPO epochs      & 5, 10 \\
					($N_{Pmb}$) PPO minibatch size       & 64, 128, 256 \\
					\bottomrule
				\end{tabular}
			\end{sc}
		\end{small}
	\end{center}
	\vskip -0.1in
\end{table}

\vspace{44pt}

\begin{table}[h!]
	
	\label{tab3.1}
	\vskip 0.15in
	\begin{center}
		\parbox{11cm}{\caption{The combination of parameters that turned out best for each experiment involving SILEM and an Ant-based body. We only ran SILEM using the Ant0 expert.}}
		\vspace{11pt}
		
		\begin{small}
			\begin{sc}
				\begin{tabular}{lcccccc}
					\toprule
					Ant-based body & $n$ & $N_D$ & $N_G$ & $d$ & $N_P$ & $N_{Pmb}$ \\
					\midrule
					Ant0    & 20 & 10 & 10 & 1 & 10 & 128\\
					Ant1 & 10 & 10 & 5 & 1 & 10 & 128\\
					Ant2    & 20  & 10 & 10 & 1 & 10 & 128\\
					Ant3    & 10 & 5  & 5 & 1 & 10 & 128\\
					Ant4      & 10  & 10 & 10 & 0.999 & 10 & 64\\
					Ant5       & 10 & 5 & 5 & 1 & 10 & 128 \\
					Ant6       & 20 & 5 & 5 & 1 & 10 & 256 \\
					\bottomrule
				\end{tabular}
			\end{sc}
		\end{small}
	\end{center}
	\vskip -0.1in
\end{table}

\newpage

\begin{table}[h!]
	
	\label{tab3.2}
	\vskip 0.15in
	\begin{center}
		\parbox{11cm}{\caption{The combination of parameters that turned out best for each experiment involving SILEM$^-$ and an Ant-based body, when the Ant0 expert is used.}}
		\vspace{11pt}
		
		\begin{small}
			\begin{sc}
				\begin{tabular}{lcccc}
					\toprule
					Ant-based body & $n$ & $N_D$ & $N_P$ & $N_{Pmb}$ \\
					\midrule
					Ant0    & 10 & 10 & 10 & 256\\
					Ant1 & 10 & 10 & 10 & 128\\
					Ant2    & 20  & 10 & 5 & 64\\
					Ant3    & 20 & 10 & 10 & 256\\
					Ant4      & 20 & 5 & 10 & 256\\
					Ant5       & 20 & 5 & 5 & 128 \\
					Ant6       & 10 & 5 & 5 & 256 \\
					\bottomrule
				\end{tabular}
			\end{sc}
		\end{small}
	\end{center}
	\vskip -0.1in
\end{table}

\begin{table}[h!]
	
	\label{tab3.3}
	\vskip 0.15in
	\begin{center}
		\parbox{11cm}{\caption{The combination of parameters that turned out best for each experiment involving SILEM$^-$ and an Ant-based body, when the learner and expert have the same Ant-based body.}}
		\vspace{11pt}
		
		\begin{small}
			\begin{sc}
				\begin{tabular}{lcccc}
					\toprule
					Ant-based body & $n$ & $N_D$ & $N_P$ & $N_{Pmb}$ \\
					\midrule
					Ant0    & 20 & 10 & 10 & 128\\
					Ant1 & 10 & 10 & 10 & 256\\
					Ant2    & 20  & 10 & 10 & 128\\
					Ant3    & 10 & 10 & 5 & 128\\
					Ant4      & 20 & 10 & 5 & 64\\
					Ant5       & 10 & 10 & 10 & 64 \\
					Ant6       & 10 & 10 & 10 & 256 \\
					\bottomrule
				\end{tabular}
			\end{sc}
		\end{small}
	\end{center}
	\vskip -0.1in
\end{table}

\section{Neural Architectures}

All of the neural networks used in this work are multi-layer perceptrons (MLPs) made up of tanh nonlinearities. For the experiments involving humanoid agents, the discriminators had 2 hidden layers with 128 units each while the policies had 2 hidden layers with 256 units each. For all other experiments (involving the Pybullet environments), the discriminators had 3 hidden layers with 128 units each, while the policies had 2 layers with 64 units each.

\section{Computing Infrastructure and Run-times}

We run our experiments on the CPUs in a computing cluster managed by \href{https://research.cs.wisc.edu/htcondor/}{HTCondor}. Each node in the cluster had approximately the following server model with minor variations: Dell PowerEdge M620, and processor: 2x Xeon E5-2670 (8 core) @ 2.60GHz. Each experimental run of SILEM takes approximately 2 days in this setup. Running the grid searches for the Ant-based bodies took approximately a week, while for the humanoid agents, it took around 2-4 days. Note that this timing is dependent on the load of the cluster during job submission. Our source code is based on Open AI baselines and the DeepMind Control Suite (for the humanoid agents), and used the following GitHub repo --- \texttt{\href{https://github.com/ywchao/merel-mocap-gail}{github.com/ywchao/merel-mocap-gail}} --- as a starting point. Our source code is attached to the supplementary materials. While SILEM incurs additional computational complexity due to the training steps for the state transformer (as compared to SILEM$^-$), we found that it did not affect training time much. The computational cost of simulating the policy's interaction with the environment is overwhelmingly higher than the other steps in an iteration of SILEM.

\bibliographystyle{named}
\bibliography{ijcai21}

%% file: fig1.tikzstyles
\usetikzlibrary{arrows.meta}

\tikzstyle{default box}=[fill={rgb,255: red,108; green,172; blue,255}, draw=black, shape=rectangle, inner sep=8pt,rounded corners=0.2cm, line width=0.2mm]
\tikzstyle{st box}=[fill={rgb,255: red,138; green,228; blue,110}, draw=black, shape=rectangle, inner sep=8pt,rounded corners=0.2cm, line width=0.2mm]

\tikzstyle{Arrow}=[-stealth, thick]
\tikzstyle{ArrowD}=[-{Triangle[scale=2.5, length=4, width=6]}, line width=2mm, dashed, color=gray]

%% file: figs/fig1a.tikz
\begin{tikzpicture}
	\begin{pgfonlayer}{nodelayer}
		\node [style=default box, font=\relsize{0.3}] (0) at (-4.25, 3.75) {Expert Trajectories};
		\node [style=default box, font=\relsize{0.3}] (1) at (2.4, 3.75) {Policy};
		\node [style=default box, font=\relsize{0.3}] (2) at (-4.25, -5.6) {Discriminator};
		\node [style=default box, align=center, font=\relsize{0.3}] (00) at (-4.25, -2.6) {Skeletal Feature \\ Abstraction};
		\node [style=default box, align=center, font=\relsize{0.3}] (11) at (2.4, -2.6) {Skeletal Feature \\ Abstraction};
		\node [style=st box, align=center, font=\relsize{0.3}] (3) at (2.4, -5.6) {Sequential \\ Affine Transform};
	\end{pgfonlayer}
	\begin{pgfonlayer}{edgelayer}
		\draw [style=Arrow] (0) -- (00) node[pos=0.45, align=center,fill=white, text width=50pt] {\includegraphics[width=1.5cm]{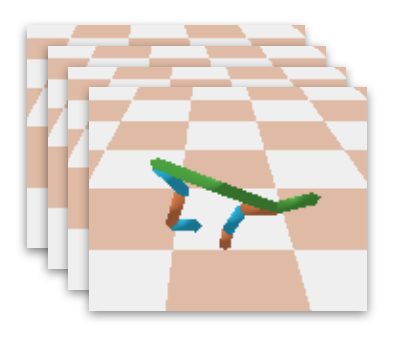}};
		\draw [style=Arrow] (00) -- (2);
		\draw [style=Arrow] (1) -- (11) node[pos=0.45, align=center,fill=white, text width=50pt, shift={(-6mm,-1mm)}] {\includegraphics[width=3cm]{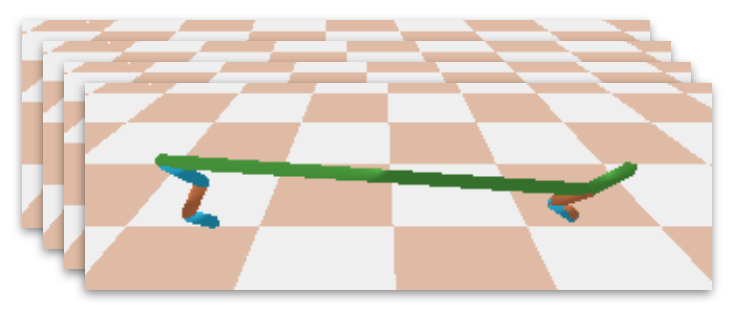}};
		\draw [style=Arrow] (11) -- (3);
		\draw [style=Arrow] (3) -- (2);
		\draw [style=ArrowD, rounded corners=5pt, font=\relsize{0.3}] (2) |- node[pos=0.75, align=center,fill=white, draw=gray, inner sep=7pt, text=black, line width=0.4mm, solid, rounded corners=5pt] {use output of Discriminator \\ as reward to train policy} ([shift={(80mm,-25mm)}]2.south east) -- ([shift={(23mm,0mm)}]1.east) -- (1);
	\end{pgfonlayer}
\end{tikzpicture}

%% file: figs/halfcheetah.tikz
\begin{tikzpicture}
	\node[label={[shift={(0mm, -10mm)}, fill=white, rounded corners=5pt, inner sep=5pt]HC0}] (A) {\includegraphics[width=4cm]{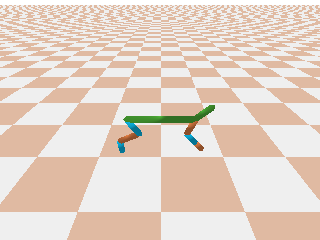}};
	\node[right=of A,label={[shift={(0mm, -10mm)}, fill=white, rounded corners=5pt, inner sep=5pt]HC2}, shift={(-5mm, 0mm)}] (C) {\includegraphics[width=4cm]{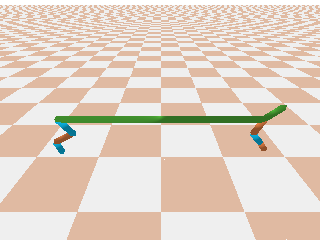}};
	
\end{tikzpicture}

%% file: figs/ant.tikz
\begin{tikzpicture}
	\node[label={[shift={(0mm, -10mm)}, fill=white, rounded corners=5pt, inner sep=5pt]Ant0}] (A) {\includegraphics[width=4cm]{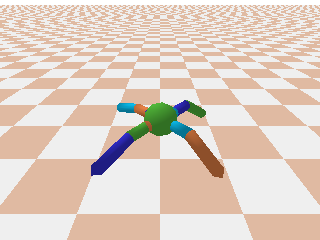}};
	\node[right=of A,label={[shift={(0mm, -10mm)}, fill=white, rounded corners=5pt, inner sep=5pt]Ant6}, shift={(-5mm, 0mm)}] (C) {\includegraphics[width=4cm]{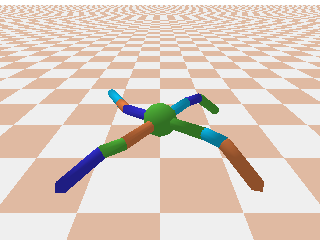}};
	
\end{tikzpicture}

%% file: figs/humandemo.tikz
\begin{tikzpicture}
	\node[label={90:CMU Mocap file}] (A) {\includegraphics[width=4cm]{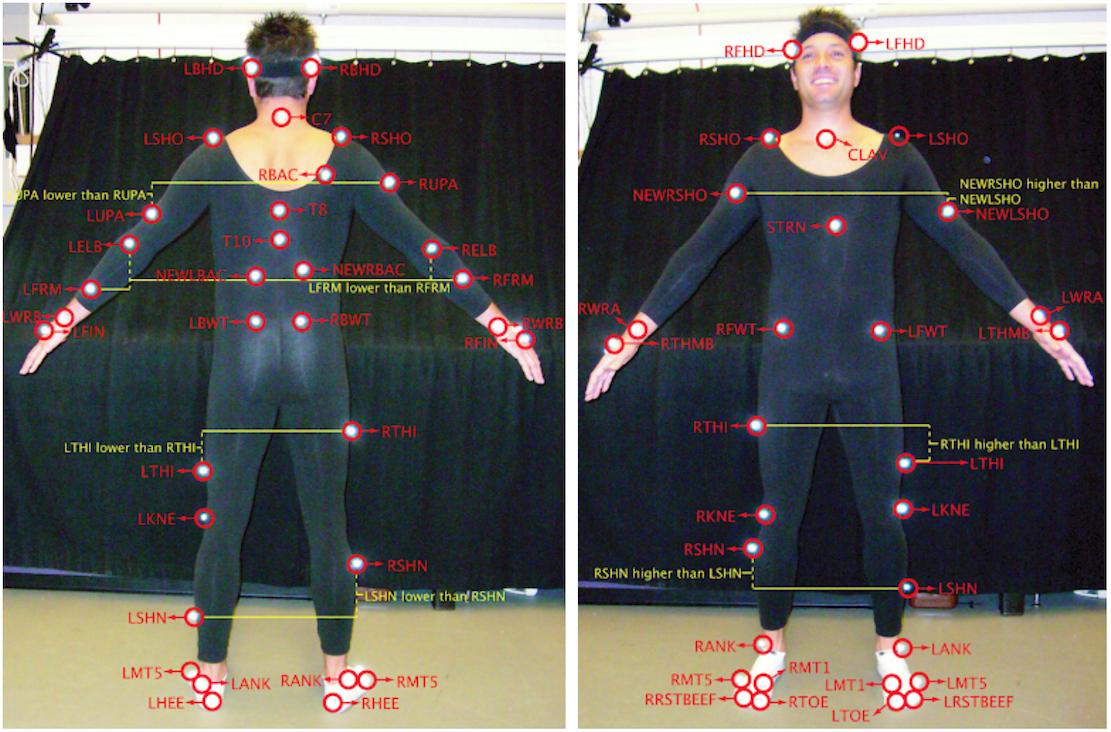}};
	\node[below=of A,label={270:Humanoid}, shift={(-35mm, 0mm)}] (B) {\includegraphics[width=2.75cm]{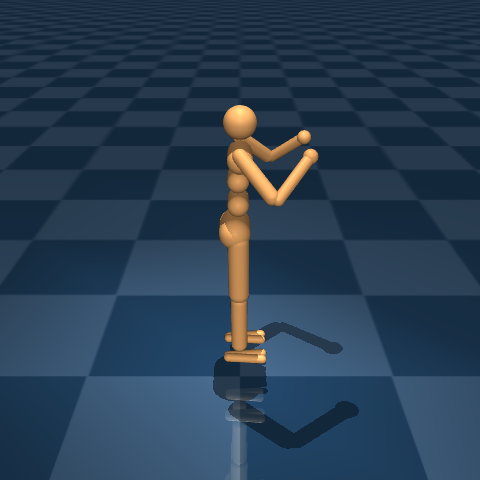}};
	\node[below=of A,label={270:HumanoidR}, shift={(0mm, 0mm)}] (C) {\includegraphics[width=2.75cm]{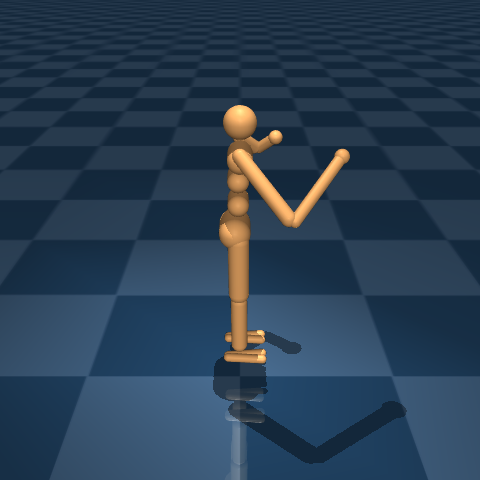}};
	\node[below=of A,label={270:Atlas}, shift={(35mm, 0mm)}] (D) {\includegraphics[width=2.75cm]{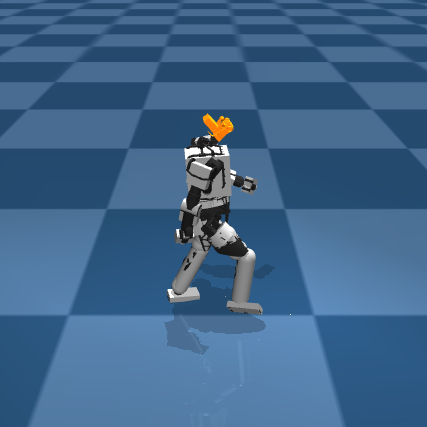}};

	\draw [style=Arrow, rounded corners=5pt] (A) -- ([shift={(0mm,-10mm)}]A.south) -| (D);
	\draw [style=Arrow, rounded corners=5pt] (A) -- ([shift={(0mm,-10mm)}]A.south) -| (B);
	\draw [style=Arrow, rounded corners=5pt] (A) -- (C);
	
\end{tikzpicture}